\documentclass{article}

\usepackage{arxiv}

\usepackage[utf8]{inputenc} 
\usepackage[T1]{fontenc}    
\usepackage{hyperref}       
\usepackage{url}            
\usepackage{booktabs}       
\usepackage{amsfonts}       
\usepackage{nicefrac}       
\usepackage{microtype}      
\usepackage{lipsum}
\usepackage{graphicx}
\usepackage{authblk}
\graphicspath{ {./images/} }
\usepackage[
backend=biber,
style=apa,
sorting=ynt
]{biblatex}

\title{Interpretability of a Deep Learning Model in the Application of Cardiac MRI Segmentation with an ACDC Challenge Dataset}

\author[1]{Adrianna Janik}
\author[1]{Jonathan Dodd}
\author[2]{Georgiana Ifrim}
\author[3]{Kris Sankaran}
\author[1]{Kathleen Curran}
\affil[1]{School of Medicine, University College Dublin}
\affil[2]{School of Computer Science, University College Dublin}
\affil[3]{Department of Statistics, University of Wisconsin–Madison}

\begin{document}
\maketitle
\begin{large}Pre-published version, official version published in Proceedings Volume 11596, Medical Imaging 2021: Image Processing; 1159636 (2021), please cite: \cite{10.1117/12.2582227} \newline
\end{large}
\begin{abstract}
Cardiac Magnetic Resonance (CMR) is the most effective tool for the assessment and diagnosis of a heart condition, which malfunction is the world’s leading cause of death. Software tools leveraging Artificial Intelligence already enhance radiologists and cardiologists in heart condition assessment but their lack of transparency is a problem. This project investigates if it is possible to discover concepts representative for different cardiac conditions from the deep network trained to segment crdiac structures: Left Ventricle (LV), Right Ventricle (RV) and Myocardium (MYO), using explainability methods that enhances classification system by providing the score-based values of qualitative concepts, along with the key performance metrics. With introduction of a need of explanations in GDPR explainability of AI systems is necessary. This study applies Discovering and Testing with Concept Activation Vectors (D-TCAV), an interpretaibilty method to extract underlying features important for cardiac disease diagnosis from MRI data. The method provides a quantitative notion of concept importance for disease classified. In previous studies, the base method is applied to the classification of cardiac disease and provides clinically meaningful explanations for the predictions of a black-box deep learning classifier. This study applies a method extending TCAV with a Discovering phase (D-TCAV) to cardiac MRI analysis. The advantage of the D-TCAV method over the base method is that it is user-independent. The contribution of this study is a novel application of the explainability method D-TCAV for cardiac MRI anlysis. D-TCAV  provides a shorter pre-processing time for clinicians than the base method.
\end{abstract}

\keywords{explainable ML, semantic segmentation, deep learning}

\section{INTRODUCTION}
\label{sec:intro} 

Cardiovascular disease is the most common cause of death worldwide \cite{who_causes_of_death}. Cardiac Magnetic Resonance (MRI) is considered the non-invasive gold standard to assess cardiomyopathy \cite{salerno_recent_2017}. Incorporating its assessment into clinical work-flow has become an important element in the investigative cardiomyopathy paradigm.  Since 2015 when deep learning technology was introduced into \cite{lecun_deep_2015} medical imaging, as well as other computer vision applications, major breakthroughs in automating analysis that were previously performed manually have become possible. With new regulations regarding decision-making systems e.g. GDPR in Europe, monitoring of these systems for quality has become increasingly important. An area causing major interest as a method for achieving higher quality in medical imaging is in the area of explainable machine learning, which is a possible direction in which new regulations could be enforced. Between the years 2017-2020 there was an especially significant rise of new explainability methods and an effort was made in this area in defining explainability in the context of machine learning  \cite{doshi-velez_towards_2017, guidotti_survey_2018, miller_explanation_2017}. Two main directions are possible with explainability methods, global and local explainability. Local explainability provides visual or textual artefacts for a single prediction, while global explainability strives to provide an understanding of the whole model. A method that provides global explainability is, for example, Testing with Concept Activation Vectors (TCAV). Producing quantification of expert-defined concepts and testing statistically for their significance is a robust method of explainability within decision-making systems. Autonomous diagnosis, however, is still a longer-term ambition in terms of being realised in daily clinical routine due to many factors that have to be taken into account and human-level expertise required in the assessment of heart condition. While full fledged decision-making systems are not yet possible, sub-tasks that lead to the diagnosis are feasible. One such sub-task is segmentation of the main structures of the heart where state-of-the-art machine learning achieves a human-expert-level of performance within a controlled environment \cite{bernard_olivier_et_al._deep_nodate}. Learning from the high-performing models by applying explainability methods may provide ueful insights about the task of segmenting cardiac structures, which could lead to a better foundation for automatic cardiac diagnosis. 
An extension of TCAV, the aforementioned explainability method includes a discovering phase (D-TCAV) and can be applied to uncover underlying features of cardiac disease from cardiac MRI data. These methods were originally designed for classification tasks. This paper considers the application of segmentation of cardiac structures (sometimes also known as pixel-level classification as each pixel is labeled). From the perspective of computer vision the task of image segmentation is different from the task of classification regarding the inputs and outputs of the models designed for these tasks. For classification the output size of the network is the same as the input size because every pixel is being classified. Whereas for segmentation, the image as a unit receives a single class label. The D-TCAV method was applied to the network that performs segmentation. The network was trained on open-source cardiac MRI data \cite{bernard_olivier_et_al._deep_nodate}, which includes patients with four different cardiomyopathies and one healthy control group. This research investigates whether concepts discovered by the network are representative for the following cardiac structures: Left Ventricle (LV), Right Ventricle (RV) and Myocardium (MYO), with respect to different
cardiac conditions present in the data the network was trained on.

\section{Related Work}
While automatic diagnosis is not currently independently possible, the research in this area can provide insight for other more realistic applications to support the diagnosis. Explainability methods mostly target explaining machine learning classifiers, which in the medical domain translate to automatic diagnostic systems. Some of these explainability methods like LIME \cite{ribeiro_why_2016}, Saliency Maps \cite{simonyan_deep_2013}, etc. could be adapted for medical applications that are currently achieving human-expert-level prediction, for example cardiac segmentation, that potentially could provide pathological insights that are beyond human-interpretation. Explainability methods applied to models achieving high performance on cardiac segmentation tasks can extract knowledge from these models about the nature of cardiac diseases.
One of the explainability methods that is applicable to semantic segmentation tasks (associating each pixel of an image with a class label e.g. LV/RV/Myocardium) is TCAV which with modifications can be a useful method of inspecting learned representation of cardiac data. The TCAV method \cite{kim_interpretability_2017} provides explanations to the users in terms of quantified user-chosen concepts corresponding to the domain-application knowledge in the classification task. Along with the accuracy of prediction, one can get supporting concepts for this prediction. An example of a relevant medical application of TCAV is the case of diabetic retinopathy (DR) diagnosis, where the method provides clinicians with a quantification of the influence different biomarkers have on the automatic diagnosis. A further, successful study in the medical domain by Clough et al. \cite{clough_global_2019} is the application of the TCAV method for cardiac disease classification. The parallel study to the one conducted by Clough et al. \cite{clough_global_2019} from the computer science perspective by Ghorbani et al. \cite{ghorbani_automating_2019} introduced an alternative procedure of obtaining concepts for TCAV called D-TCAV, where the user is no longer required to provide concepts manually but is presented with relevant concepts learned by the network. D-TCAV builds on the work on TCAV that has been applied successfully in the medical domain and hence D-TCAV is a promising candidate method to be tested in the medical domain. The use of the D-TCAV method for discovering concepts in cardiomyopathies may provide clinicians with previously unknown aspects of these pathologies that can be detected automatically.

\section{Methods}
\subsection{Data}
The dataset utilised in this study was a set of cardiac MRI images coming from the open-source challenge MICCAI 2017 \cite{bernard_olivier_et_al._deep_nodate}. The dataset is divided into five evenly distributed subgroups - one healthy control group and four cardiomyopathies: chronic myocardial infarction, dilated cardiomyopathy, hypertrophic cardiomyopathy and dilated right ventricle. The training dataset includes regions of interest labeled by clinical experts and includes data from 70 patients in the training phase and 30 in the development phase. A single scan includes sevral views of the heart, the key view being the short axis, which contains several slices (between 6 and 18) of the heart from the base to the apex. Each slice is obtained by acquiring a 2D cine-MRI steady-state-free-precesion sequence with retrospective ECG-gating. Each slice has a thickness of 5-8 mm and some datasets include an inter-slice gap of 5 mm. The MRI scans were acquired over a 6 year period either on a 1.5 T MRI (Siemens Aria, Siemens Medical Solutions, Germany) or 3.0 T (Siemens Trio Tim, Siemens Medical Solutions, Germany). The spatial resolution ranges from 1.37 to 1.68 mm2/pixel and 28 to 40 images cover completely the cardiac cycle.

\begin{figure}
    \centering
    \includegraphics[width=0.51\textwidth]{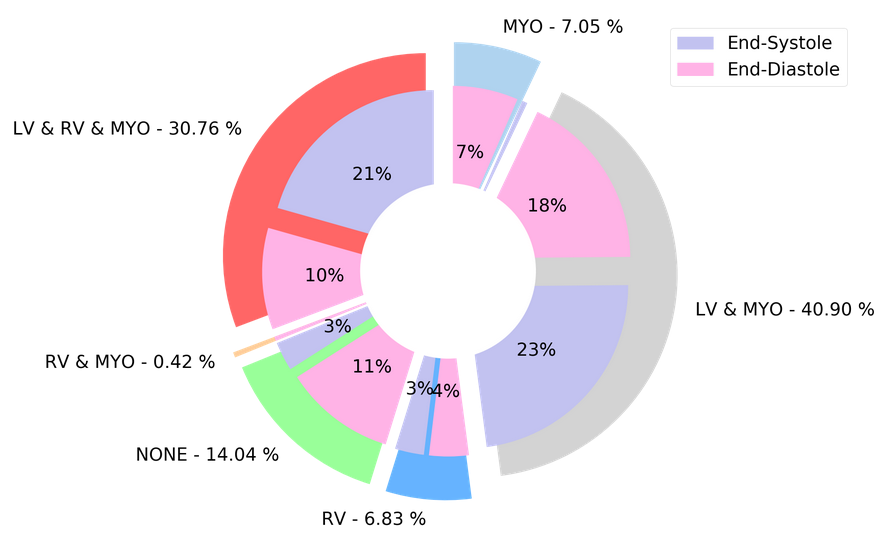}
    \caption{Distribution of data according to structures present in the image and phase (end-systole, end-diastole). There are only ~31\% of images that contain all 3 structures of interests, on the other hand, there are ~14\% that contain none of them.}
    \label{fig:dist}
\end{figure}

The data samples can be characterized in terms of presence or absence of cardiac structures in the view, see Figure \ref{fig:dist}. Data exploration of this dataset showed that the following imaging subgroups exist in the dataset: None of the structures are present (NONE). All of the structures are present (Left Ventricle \& Right Ventricle \& Myocardium; LV \& RV \& MYO). Only myocardium is present (MYO). Only the right ventricle is present (RV). Both the right ventricle and myocardium are present (RV \& MYO). Both the left ventricle and myocardium are present (LV \& MYO).

The number of samples per imaging subgroup according to prevalence are shown in Figure \ref{fig:dist}. The sample in this example is one slice of an MRI dataset, and typically, one patient scan contains an average of 10 slices of the heart in diastole and systole. Such large datasets make the task of cardiac segmentation using machine learning methods challenging. 
Figure \ref{fig:dist} also shows that the distribution of classes across images vary, which makes the evaluation of the system trained on such data not trivial even with the assumption that there are no ground truth errors.

Training Machine Learning models involves an optimisation step that minimizes the chosen metric measuring error. With the assumption that the lower that error is the better the model. The metrics that utilise overlapping regions of the detected structures with the original mask (e.g. DICE score) are prone to bias \cite{metrics}. An example would be images where there are no relevant cardiac structures to detect and are therefore receiving low overlap scores and the network is being penalized for this reason during the training phase, in other words it is being penalized for a desired behaviour. Despite this fact in this study we report DICE score as it is a standard score for assessing semantic segmentation models.

{}

\subsection{Discovering and Testing Concept Activation Vectors}\label{sec:dtcav}

\begin{figure}
    \centering
    \includegraphics[width=0.5\textwidth]{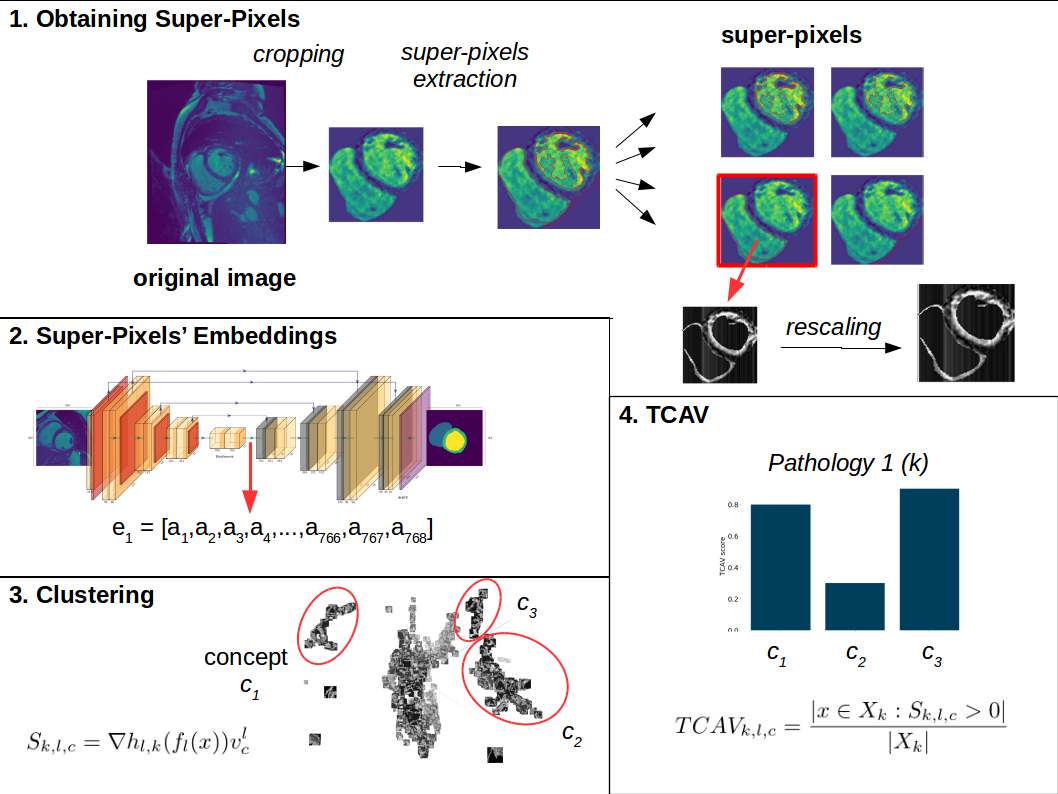}
    \caption{Illustration of the D-TCAV method, consisting of four steps, applied to cardiac data. The first step is to crop cardiac images to the region of interest centered around the heart and extract super-pixels from obtained patches by clustering similar pixels together using  the SLIC algorithm and then re-scale to original data size. In the second step super-pixels are input into the deep network (U-Net) to collect activations in the middle layer (embeddings). The third step requires running clustering on the latent space (embeddings) of super-pixels and cluster selection, which are the concepts tested in the final step with TCAV.}
    \label{fig:flowchart}
\end{figure}

The D-TCAV method consists of four main steps (Figure \ref{fig:flowchart}). The first step is to perform multi-resolution segmentation of images, which means fragmenting the same image several times with different sizes of these fragments. To do that D-TCAV applies the SLIC (Simple Linear Iterative Clustering) method \cite{achanta_slic_2012}. SLIC is an algorithm that creates clusters of pixels with similar properties called super-pixels. The clustering is based on signal similarity and location proximity. In this work we operate only on pixel intensity values. Super-pixels are assumed to display perceptual meaning since pixels belonging to a given super-pixel share similar visual properties. From one image, up to 5 super-pixels are obtained.
This is followed by the second step to resize segments to the original input size of the network (348x348). Resized patches are then processed by the network and activation signals from the middle layer are collected (Figure \ref{fig:flowchart}, step 2). The process of discovering concepts is carried out by the third step of clustering the latent space, that is a representation of compressed data in which similar data points are closer together in space, for every class with outliers removal. Clusters are then compressed to Concept Activation Vectors (CAVs) defined as a perpendicular vector to the decision boundary between cluster data and random counterparts obtained by training a binary classifier.
The final step of testing is performed with the usage of the original TCAV score as presented by Kim et al. \cite{kim_interpretability_2017}. TCAV score is defined as the count of positive values of directional derivatives between CAV and gradients of layer from which the CAV was obtained and normalised to the size of class for which TCAV is calculated, this is further explained in Equation \ref{eq:tcav}. The result is the set of concepts ranked by the importance score per each class. Figure \ref{fig:flowchart} illustrates the D-TCAV method as applied in this study consisting of four steps applied to cardiac MRI data, the details of TCAV method between step 3 and 4 were omitted for clarity and only equations are recalled there.

\subsection{Segmentation Model}\label{sec:model}
The architecture utilised in this project was a U-Net network as described by Ronneberger et al. \cite{ronneberger_u-net:_2015}. The goal was to recreate the state-of-the-art performance achieved by networks presented by Isensee et al. \cite{isensee_automatic_2018} and Baumgartner et al. \cite{baumgartner_exploration_2017} for the MICCAI ACDC 2017 challenge \cite{bernard_olivier_et_al._deep_nodate}. The main differences between networks are as follows: Isensee et al. \cite{isensee_automatic_2018} trained an ensemble of networks achieving the highest scores in the benchmark dataset. This requires substantial computational resources, and this work follows the approach by Baumgartner to focus on the 2D network as 3D networks showed no significant improvements, while still the network achieved state-of-the-art performance on the task of segmenting the left ventricle, right ventricle and the myocardium.
Two models were trained, one with and one without transformations, and the performance is presented in table I. In the table performance of two models is reported. The first one performance on training and development datasets constituting respectively 70 and 30 cases for a simple U-Net without any augmentations. The second model was trained on data transformed with the following augmentations: elastic transform, rotation, scaling, with probability 10\%.
Deep neural network U-net was trained with the parameters presented as follows. U-Net was initialized with 48 filters and the number of filters were doubled with each step on the encoding path, the model was trained with a batch size of 10 for 175 epochs. To increase the size of the dataset (for better results) augmentation technique were used, images were re-scaled to the size of 348x348 and were augmented at random with a probability of 0.1 with scaling transformations where scale range was randomly sampled from interval 0.75 and 1.25, elastic transformation and rotation.

  \begin{table}[]\label{tab:model-tabel}
  \begin{center}   
\begin{tabular}{l|l|c|c|c|c|c|c|c}
            \multicolumn{2}{c|}{}&  \multicolumn{7}{c}{Dice Score}             \\
            \hline
           \multicolumn{2}{l|}{model}& \multicolumn{2}{c|}{LV} & \multicolumn{2}{c|}{RV} & \multicolumn{2}{c|}{MYO} & global      \\
            \hline
             &dataset & avg   & median & avg   & median & avg   & median & avg \\
            \hline
1 & train & 93  & 100  & 83  & 100  & 93  & 100  & 91      \\
& dev   & 69  & 81  & 43  & 46  & 64  & \textbf{81}  & \textbf{61}      \\
\hline
2 & train & 92  & 98  & 80  & 95  & 89  & 94  & 88      \\
&dev   & 72  & 86  & 53  & 82  & 69  & \textbf{82}  & \textbf{66}      \\
\end{tabular}
\caption{Models performance. In the table performance of two models is reported. The first one performance on training and development datasets constituting respectively 70 and 30 cases for a simple U-Net without any augmentations. The second model was trained on data transformed with the following augmentations: elastic transform, rotation, scaling, with probability 10\%.}
\end{center}
 \end{table}
 
\subsection{Cardiac D-TCAV}
This study applies the D-TCAV method to cardiac data. This section presents the application specific aspects and details of the approach utilised.
Cardiac semantic segmentation, in this study, is formulated as learning the function of input X (a deep learning model):  f(X) = Y, where X is an input image represented as a 2 dimensional matrix of size n x n containing pixels intensities, and Y is an output matrix of the same size containing masked pixels. Every pixel in Y has a value $c \in \{MYO, RV, LV, bg\}$ encoding the class the pixel belongs to. 
Data input to the network was a single MRI slice from a single subject. The hierarchical structure of different slices and views was not utilised at the training stage. The segmentation model was implemented and trained according to the description in Section \ref{sec:model}. In order to collect meaningful latent representation appropriate functions had to be implemented to facilitate user access to the middle layer of the network. This was required in order to access the activation signals from the middle layer, calculate gradients with respect to the middle layer, and save the context information that could be further used to analyze results. The discovering phase of D-TCAV method required that the input data was fragmented into super-pixels using the SLIC method. The number of slices was selected based on manual adjustments and qualitatively five super-pixels were selected for this step. The slices had such characteristics that only  $\sim$10\% of pixels in the image belong to the heart ( $\sim$8\% in end-systole and $\sim$12\% in end-diastole). To increase this ratio the cropping approach was implemented before images were further processed with the SLIC method. The region of interest (ROI) was obtained by utilising ground-truth masks and cropping original images according to them. Once input data was processed with the cropping phase, data was ready to be transformed with the trained network. The U-Net model has an encoder-decoder structure that was trained in this setting. The data is entered to the network and the signal is collected prematurely at the middle layer after the encoding phase (see step 2 in Fig. \ref{fig:flowchart}), resulting in outputting the latent space representation of the image. This latent representation was then further analysed with unsupervised machine learning methods. Firstly, UMAP dimensionality reduction method \cite{mcinnes_umap_2018} was selected and the data was reduced with this method, this dimensionality reduction method was chosen as one that gives good approximation for non-linear data. Secondly, this new representation was clustered with the k-means clustering method as used in the original article and also because other methods that were tried did not prove to be better than k-means. Thirdly, clusters were reviewed and outliers removed. Finally, from among the identified clusters concepts were selected to further train Concept Activation Vectors. The latent representation was also obtained for the training data, also with additional information - gradients were calculated with respect to the middle layer. Collected gradients along with the CAVs (introduced in Section \ref{sec:dtcav}) were finally incroporated in the calculation of TCAV scores. For each pathology from the dataset, TCAV score was calculated with respect to the concepts. The final step was to evaluate the D-TCAV approach to assess the clinical interpretation of the cardiac concepts output from the model, with the clinical expertise of a radiologist.

\section{Results \& Experiments}
\begin{figure}
    \centering
    \includegraphics[width=0.5\textwidth]{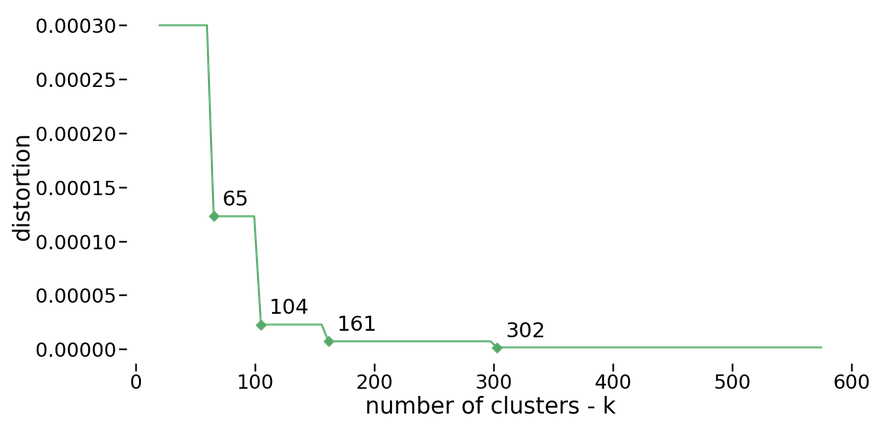}
    \caption{This graph represents the curvature of distortion of data samples, defined as the average of the squared distances from the cluster centroids using euclidean distance metric. Using this approach we can identify the number of clusters that could yield good split of data.}
    \label{fig:distortion}
\end{figure}

 The model was trained as decribed in Section \ref{sec:model}. Interpretation of the results of segmenting multi-slice images across time is not trivial \cite{metrics}. Discrepancy between the median score and the mean score suggests that the network did not handle the outliers well.
The latent space of super-pixels activations was clustered using the k-means algorithm following the original D-TCAV implementation. The number of clusters was heuristically determined by assessing the curvature of the distortion of data samples Figure \ref{fig:distortion} – defined as the average of the squared distances from the cluster centroids using the euclidean distance metric. The number of clusters with the highest curvature was chosen as the optimal number of clusters, this method is sometimes referred to as an elbow method. After applying a k-means algorithm with 65 clusters the smallest cluster size is 5, the largest: 511, the average: 237.8 and the median: 222.0.
Initial qualitative evaluation of selected clusters showed that several clusters could be considered as representations of certain cardiac properties which is promising. Example images from one of these clusters are presented in Figures \ref{fig:56cls}. The experiment focused on the best strategy for cluster selection and assessment of the quantitative value of each concept.

\subsection{Concept Selection Strategy}
The strategy employed in this work to select concepts was inspired by the work in D-TCAV, where concepts were selected based on their sizes and variability. Cardiac MRI images, however, are very different from images used in the original work, which contained images representing non-medical data. Our highly specialized medical dataset required a new approach to concept selection. The range-based \cite{ghorbani_automating_2019} selection was utilised with adapted ranges based on the count of the patches coming from a specific class. Then we propose to compare this approach with clusters not selected as the concepts, not using pre-selection at all and treating each cluster as a concept to avoid omitting clusters that could potentially yield interesting results. From 18 clusters that were selected as concepts during the pre-selection phase only 8 yielded corresponding TCAV scores, others were either not statistically significant or the CAVs were not available due to all directional derivatives being less than or equal to zero (Equation 1). From the clusters not identified as concepts during pre-selection another 25 clusters received TCAV scores as statistically significant. A count-based selection of clusters may not be the best method of selecting concepts if it is used solely without other support because the data has a complex structure, but it is a good method for eliminating spurious concepts like the one that consists of patches coming only from one patient but different slices.

\subsection{Concepts Quantitatively}\label{quant}
\begin{figure}
    \centering
    \includegraphics[width=0.4\textwidth]{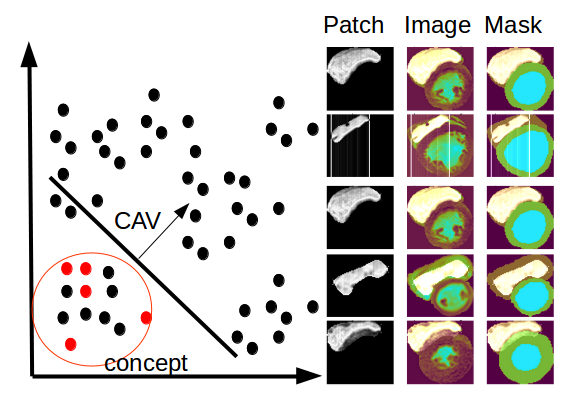}
    \caption{Overview of one of the concepts partially representing right ventricle. On the left the CAV vector is presented in the 2 dimensional projection of the latent space, dots represent patches of concepts (red the one showed on the right). On the right set of three images is presented with example patches from the concept selected. First image represent a patch, second original image with patch overlay-ed on top and the last mask with patch overlay-ed. This concept received following TCAV scores 0.48 for MINF, 0.23 for HCM and 0.26 for abnormal RV.}
    \label{fig:56cls}
\end{figure}

\begin{figure}
    \centering
    \includegraphics[width=0.4\textwidth]{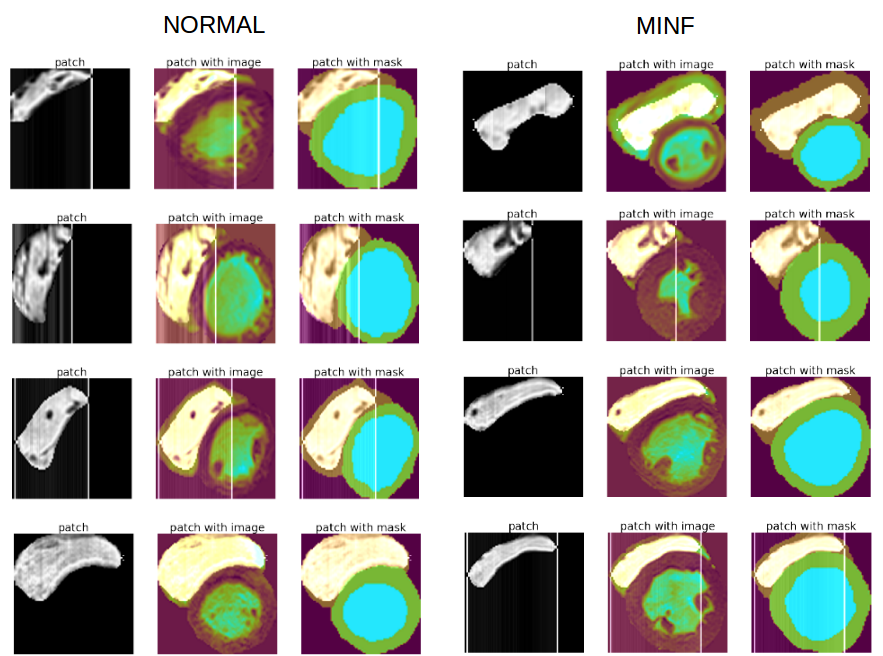}
    \caption{Examples of patches from concept data along with their overlay-ed version with original image and mask. Normal control group examples on the left for concept 56 in end-diastole. MINF examples on the right in end-diastole.}
    \label{fig:nor_minf}
\end{figure}

\begin{figure}
    \centering
    \includegraphics[width=0.4\textwidth]{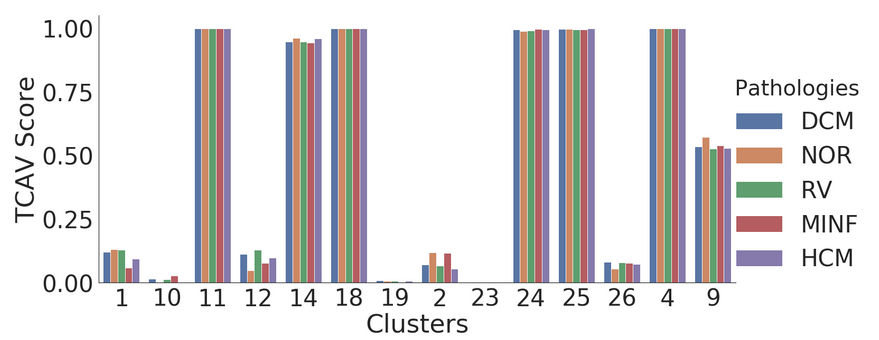}
    \caption{TCAV scores part 1. The Figure presents TCAV scores for each identified concept for which it was possible to calculate TCAV score. TCAVs are calculated for each class per each concept. The two figures \ref{fig:tcavs1} and \ref{fig:tcavs2} present TCAVs scores for all concepts.}
    \label{fig:tcavs1}
\end{figure}

\begin{figure}
    \centering
    \includegraphics[width=0.4\textwidth]{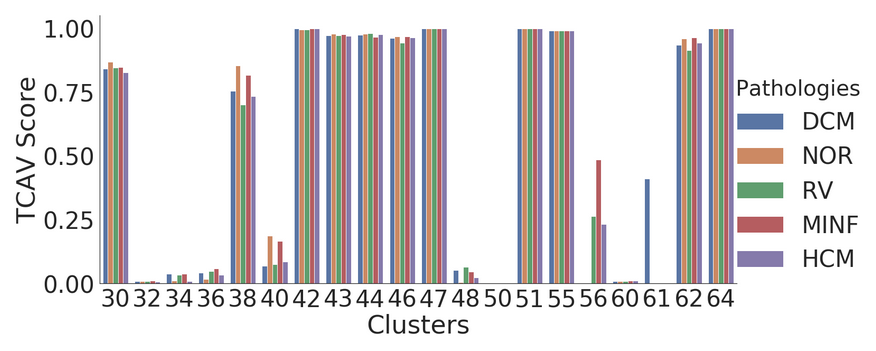}
    \caption{TCAV scores part 2. Same chart as in Figure \ref{fig:tcavs1} but for the remaining set of concepts.}
    \label{fig:tcavs2}
\end{figure}

\begin{figure}
    \centering
    \includegraphics[width=0.5\textwidth]{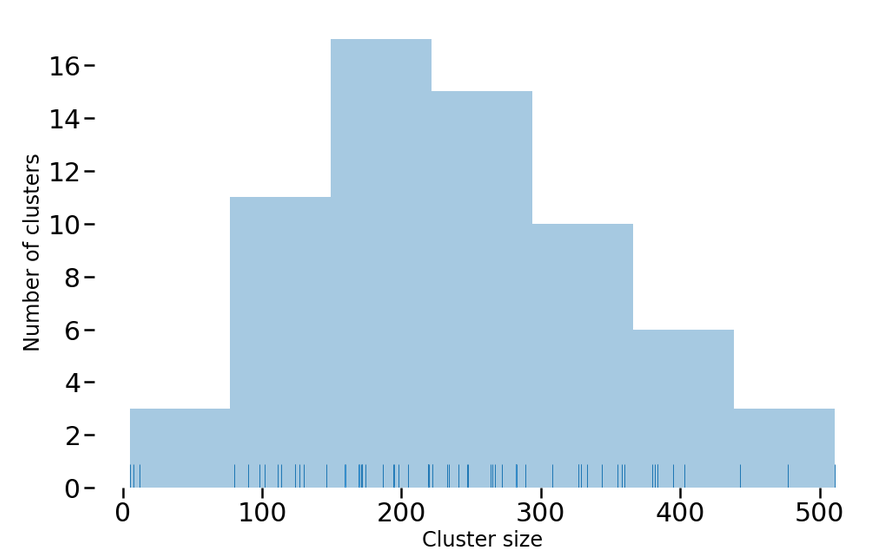}
    \caption{Figure represents the distribution of clusters' sizes, mean size of a cluster is 237.8 and median is 222.0.}
    \label{fig:clusters}
\end{figure}

\begin{figure}
    \centering
    \includegraphics[width=0.5\textwidth]{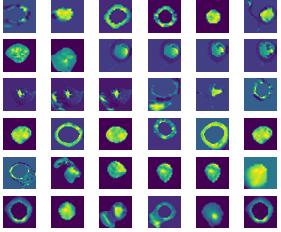}
    \caption{Cluster number 12.}
    \label{fig:cluster12}
\end{figure}

\begin{figure}
    \centering
    \includegraphics[width=0.5\textwidth]{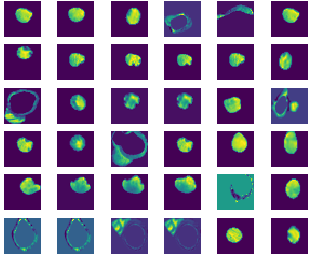}
    \caption{Cluster number 61.}
    \label{fig:cluster61}
\end{figure}

\begin{figure}
    \centering
    \includegraphics[width=0.5\textwidth]{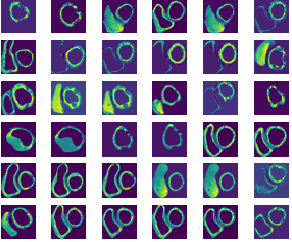}
    \caption{Cluster number 4.}
    \label{fig:cluster4}
\end{figure}

\begin{figure}
    \centering
    \includegraphics[width=0.5\textwidth]{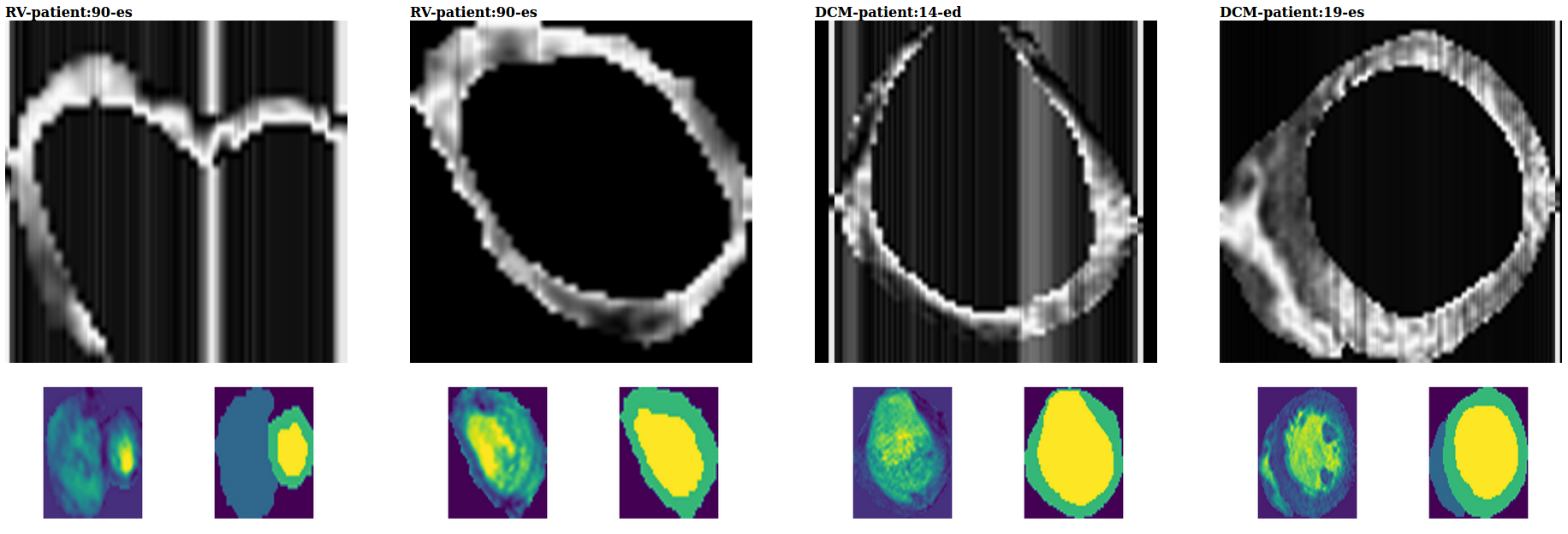}
    \caption{Cluster number 62, examples of segments obtained from SLIC with context image of heart and its mask.}
    \label{fig:cluster62}
\end{figure}

TCAV scores were obtained using the equation proposed by Kim et al. \cite{kim_interpretability_2017}, recalled below in equation \ref{eq:tcav}, that uses equation \ref{eq:dirder}.
\begin{equation}\label{eq:dirder}
    S_{k,l,c} = \nabla h_{l,k}(f_l(x)) v^l_c
\end{equation}
\begin{equation}\label{eq:tcav}
    TCAV_{k,l,c} = \frac{|x \in X_k: S_{k,l,c} > 0|}{|X_k|}
\end{equation}
The statistical significance of the scores is tested using t-test, with alpha equal to 5\%. The scores in Figure \ref{fig:tcavs1} and \ref{fig:tcavs2} are reported only for clusters where the t-test rejected the null hypothesis of the TCAV score being a random score (where the TCAV score mean was not statistically different from the TCAVs mean trained on randomly chosen CAVs on a sample size of 100 datasets). For all the clusters that had valid TCAV scores the mean score of the TCAV score for categories was calculated along with standard deviation and a percentage difference between maximum and minimum value of the score to measure how much the scores differ among each other. To see distribution of data among clusters see Figure \ref{fig:clusters}, for example patches belonging to different clusters see Figures: \ref{fig:cluster12}, \ref{fig:cluster61}, \ref{fig:cluster4}, \ref{fig:cluster62}. On average scores minimum and maximum values per cluster was different by 3.4\% with standard deviation of 5.24\%, this indicates that the scores for different pathologies and control group do not differ much among each other for the given cluster. Maximum difference between scores was 25.49\% and was yielded by the cluster 56 which received a very high score for MINF class (0.49) no scores for DCM and NOR, and 0.26 for RV and 0.23 for HCM (see Figure \ref{fig:56cls} and \ref{fig:nor_minf}).

\section{Discussion}
In this study one of the machine learning interpretability methods - D-TCAV \cite{ghorbani_automating_2019} was applied to the task of cardiac segmentation of left ventricle, right ventricle and myocardium for the cardiac MRI data of 100 patients with four groups of different cardiomyopathies and a control group \cite{bernard_olivier_et_al._deep_nodate}. D-TCAV was applied for the first time to the clinical data in the setting of semantic segmentation (pixel-level classification). Previous work on the base method TCAV was applied to detection of coronary artery disease for Biobank cardiac dataset providing explanations for the deep learning classifier predictions. D-TCAV method introduced enhancement for the TCAV method that does not require from the clinicians hand crafted datasets of images representing concepts in terms of which the explanations are constructed, reducing the time of method application and pre-processing required to craft the data. 

The explanations of the deep learning model for segmenting LV, RV and Myocardium in D-TCAV comes as a quantified subsets (clusters) of the original data.
These subsets called concepts represent fragments of original images (coming from different patients and different phases of a cardiac cycle).
There were many clusters that received statistically significant TCAV scores (Figure \ref{fig:tcavs1} and \ref{fig:tcavs2}, Section \ref{quant}), but one observation from their values is that they lie very close to each other among different cardiac conditions (also Section \ref{quant}). This may be caused by the fact that the network was trained with a different objective than classification. The deep network was trained to segment images into anatomical structures (LV, RV, Myocardium), which means that the information about each patient condition was not utilised during training, the objective was not to distinguish between the cardiomyopathies but to segment the anatomical parts (LV, RV, Myocardium) which is consistent with the results. It may suggest that the network that was trained for semantic segmentation of cardiac structures that has no abstract information on cardiac conditions provided during training is not learning cardiac condition characteristics directly. This means that specialized features have to be extracted explicitly, either by imposing this through a training objective or by hand-crafting features after semantic segmentation. This entails the need for a validation study in clinical setting and assess discovered concepts using a broad range of clinical metrics. It also may be the case that the increased number of clusters  obtained from for example hierarchical clustering could yield better results being more localized. The concepts representing cardiac parts (LV, RV, Myocarium) were identified as important in all the cardiac pathology classes, this suggests that the concepts are more representative of the heart anatomically than for underlying condition.

There are several approaches to achieve cropping from step 1 in Figure \ref{fig:flowchart} which in literature is also called automatic heart localization (e.g. \cite{sorgel_automatic_1997}).
 The cropping phase in the D-TCAV method could be approached in different ways but this is not the focus of this study. One example approach to obtain a region of interest (ROI) is to process each image with a Fourier Analysis and crop images based on their frequency distributions \cite{noauthor_17_nodate}. Another approach which was utilised in this project is to utilise ground-truth masks and crop original images according to them. This simple approach did result in ROIs with high percentage of pixels belonging to a heart anatomical structure and was suitable for the use case analysed here, but could be substituted with automated approach for unseen data.

\begin{figure}
    \centering
    \includegraphics[width=\textwidth]{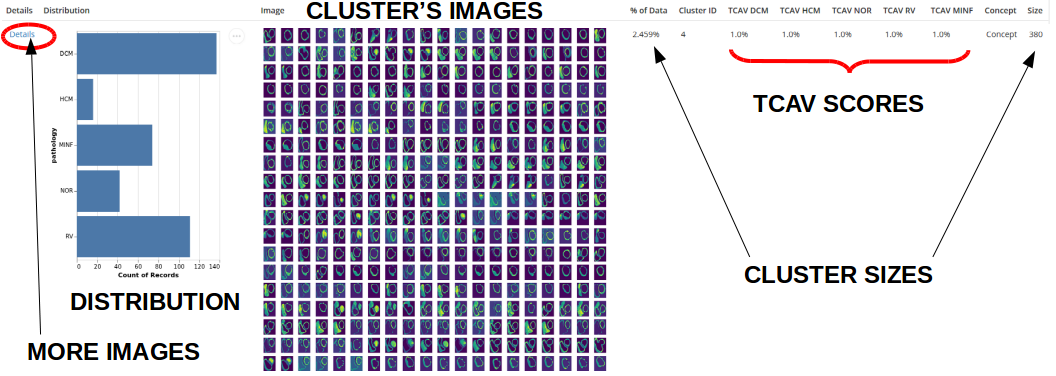}
    \caption{Sample record from the publicly available table. Online interface for the data provides a useful way of sharing results and reviewing different concepts and clusters. Along with cluster images examples, there are available the following information: cluster size (percentage and counts), TCAV scores per each class, distribution of classes among patches in the cluster and link to all data samples from the cluster in a separate page with further details.}
    \label{fig:table}
\end{figure}

The information processed in this project regarding clusters review is published online on GitHub under the following address: \url{https://adri-j.github.io/datasets/}, for an overview of the interface see Figure \ref{fig:table}. This page contains information about each cluster discovered in the data with samples of images in each cluster, cluster id, distribution of patches in the cluster across different classes (NOR, RV, MINF, DCM, HCM), size of the cluster along with the percentage it constitutes, TCAV scores for all the classes (pathology and a control group), and also an information whether the cluster was identified as a concept or not. Finally, there is a link to a subpage with all the images from the given cluster.
For the purposes of reproducible research, all the files and patches and code have been made available on GitHub \url{https://github.com/adri-j/cardiac-mri-segmentation-dtcav.git}.

To summarize, this study applied existing interpretability method D-TCAV to a new task of the semantic segmentation for the first time applied to clinical data. The resulting concepts are consistent with semantic segmentation task but needs further clinical validation. 

\clearpage

\section*{Acknowledgments} 
This work was funded by Science Foundation Ireland through the SFI Centre for Research Training in Machine Learning (18/CRT/6183).
\printbibliography 

\end{document}